\documentclass{article}

    \PassOptionsToPackage{numbers, compress}{natbib}

    \usepackage[final]{tackling_climate_workshop_style}

\usepackage[utf8]{inputenc} 
\usepackage[T1]{fontenc}    
\usepackage{hyperref}       
\usepackage{url}            
\usepackage{booktabs}       
\usepackage{amsfonts}       
\usepackage{nicefrac}       
\usepackage{microtype}      
\usepackage{graphicx}
\usepackage{amsmath}
\usepackage{float}
\usepackage{ragged2e}
\usepackage{appendix}
\usepackage{graphicx}

\title{Estimating Atmospheric Variables from Digital Typhoon Satellite Images via Conditional Denoising Diffusion Models}

\author{%
Zhangyue Ling$^{1}$ \quad Pritthijit Nath$^{2}$ \quad César Quilodrán-Casas$^{3,4,5}$\\ 
$^1$ Department of Computing, Imperial College London\\
$^2$ Department of Applied Math and Theoretical Physics,
  University of Cambridge\\
$^3$ Department of Earth Science and Engineering, Imperial College London\\ $^4$ Grantham Institute for Climate Change and the Environment, Imperial College London\\
$^5$ National Center for Artificial Intelligence CENIA, Chile\\
\texttt{\{zhangyue.ling23, c.quilodran\}@imperial.ac.uk};
\texttt{pn341@cam.ac.uk}\\
}

\begin{document}

\maketitle

\begin{abstract}
This study explores the application of diffusion models in the field of typhoons, predicting multiple ERA5 meteorological variables simultaneously from Digital Typhoon satellite images. The focus of this study is taken to be Taiwan, an area very vulnerable to typhoons. By comparing the performance of Conditional Denoising Diffusion Probability Model (CDDPM) with Convolutional Neural Networks (CNN) and Squeeze-and-Excitation Networks (SENet), results suggest that the CDDPM performs best in generating accurate and realistic meteorological data. Specifically, CDDPM achieved a PSNR of 32.807, which is approximately 7.9\% higher than CNN and 5.5\% higher than SENet. Furthermore, CDDPM recorded an RMSE of 0.032, showing a 11.1\% improvement over CNN and 8.6\% improvement over SENet. A key application of this research can be for imputation purposes in missing meteorological datasets and generate additional high-quality meteorological data using satellite images. It is hoped that the results of this analysis will enable more robust and detailed forecasting, reducing the impact of severe weather events on vulnerable regions. Code accessible at %
\url{https://github.com/TammyLing/Typhoon-forecasting}. 
\end{abstract}

\section{Introduction}
 In recent years, the frequency and intensity of extreme weather events have increased due to the impacts of global climate change. In particular, typhoons which are one of the types of tropical cyclones (TCs) get notable attention because of the harm they cause to both the natural environment as well as human societies. Taiwan, one of the major economic hubs and populous regions in the Asia, is particularly significant due to its high vulnerability to typhoons \cite{xu2015systemic}, making it an excellent focus region for this project.With the rapid development of Machine Learning, the typhoon forecasting can be done by numerous methods. Rita et al. \cite{kovordanyi2009cyclone} pioneered the use of artificial neural networks to analyze satellite image data, which marked the first application of deep learning techniques in typhoon trajectory forecasting. R{\"u}ttgers et al. \cite{ruttgers2019prediction} were the first to utilise satellite images and a generative adversarial network (GAN) to successfully predict both the center coordinates of typhoons and the future shape of cloud structures around typhoons. Moreover, diffusion models have been used for predicting tropical cyclones \cite{nath_forecasting_2023}. DYffusion \cite{cachay2023dyffusion} was one of the improved diffusion models that introduced a time-conditional interpolator and a predictor network to make a multi-step probability prediction. Building on these advancements, this project uses Conditional Denoising Diffusion Probability Models (CDDPM) \cite{saharia2022palette} to predict multiple ERA5 reanalysis variables simultaneously, using Digital Typhoon (DT) \cite{kitamoto2024digital} images as conditional inputs. CDDPM has been successfully used in remote sensing previously \cite{zhang2024inpainting, qosja2024sar, haitsiukevich2024diffusion, guan2023diffwater}

The study contributions to the current literature can be summarised as follows:
\begin{enumerate}
    \item A customised typhoon dataset is created, facilitating the matching of DT and ERA5 reanalysis data for any given region, allowing more accurate forecasting models across various regions for the future study.

    \item We successfully demonstrate CDDPM to consistently outperform other  consistently outperformed other three models across multiple reanalysis variables with the highest PSNR and SSIM scores and the lowest FID and LPIPS score.

    \item As a result we also demonstrate the ability to generate additional high-quality reanalysis data from satellite images, which can be used both for forecasting and to fill in gaps in existing reanalysis datasets. 
\end{enumerate}
 
\section{Data}
\subsection{Datasets}
\begin{itemize}
    \item ERA5 \cite{hersbach2020era5} is the fifth generation ECMWF reanalysis for the global climate from 1940 onwards provided by the European Centre for Medium-Range Weather Forecasts (ECMWF). In this project, this dataset is used as the model input data and the variables used are the u-component of wind (u10), v-component of wind (v10), surface pressure (sp), and temperature at 2 meters (t2m) within the latitude and longitude range of \(116.0794\)-\(126.0794\) and \(18.9037\)-\(28.90374\), respectively, covering an area around Taiwan ( \(\pm 5^\circ\)).
    \item Digital Typhoon Dataset \cite{kitamoto2024digital} is the largest typhoon satellite image dataset covering the Western North Pacific region from 1978 to 2022, with a high resolution of 5 km and a temporal resolution of 1 hour. As this project focuses on the area around Taiwan, 129 typhoons filtered out from original collection is used as the target dataset.
\end{itemize}

\subsection{Data Processing}
\begin{enumerate}
    \item \textbf{Region Selection and Data Extraction}: Selected Taiwan region from $18.9037^\circ \text{N}$ to $28.9037^\circ \text{N}$ and $116.0794^\circ \text{E}$ to $126.0794^\circ \text{E}$. Extracted relevant typhoon data and cropped it to ensure that the selected region of interest remains fixed.

    \item \textbf{Data Cleaning, Alignment, and Normalisation}: Cleaned the datasets by removing errors and inconsistencies. Synchronised Digital Typhoon and ERA5 data to ensure consistent use in model training and inference. Standardised each variable by normalising the data to a range of [0, 1], ensuring consistent scaling across all input features.

    \item \textbf{Data Augmentation and Train-Test Split}: Applied augmentation techniques such as random noise addition, Gaussian smoothing, and contrast adjustment to improve the model's generalisation capability. Divided the dataset into 80\% for training and 20\% for testing, with augmentation applied to the training data.
\end{enumerate}

\section{Methodology}

\subsection{Training}
In this project, the forward diffusion process involves gradually transforming the ERA5 meteorological data into pure noise by incrementally adding Gaussian noise to the input ERA5 variables (u10, v10, sp, and t2m) over $T$ time steps. Each step slightly degrades the data by blending it with a small amount of noise, eventually resulting in a nearly uniform noise distribution. This noisy data serves as the basis for training the model to reverse the process.

The reverse diffusion process aims to reconstruct the original clean reanalysis data (ERA5) from the noisy observations generated during the forward process. In the context of CDDPM, the reverse process is parameterised by a neural network \( p_\theta \). The network takes as input the conditional data \( x \) (Digital Typhoon data), the noisy observation \( y \) (Noisy ERA5), the noise level $\gamma$, and predicts the noise component $\epsilon$. The goal is made to to minimize the difference between the predicted and true noise, thereby learning to effectively reverse the diffusion process.

\begin{figure}[!h]
    \centering
    \includegraphics[trim = {0cm 0cm 0cm 0cm}, clip,width=0.75\hsize]{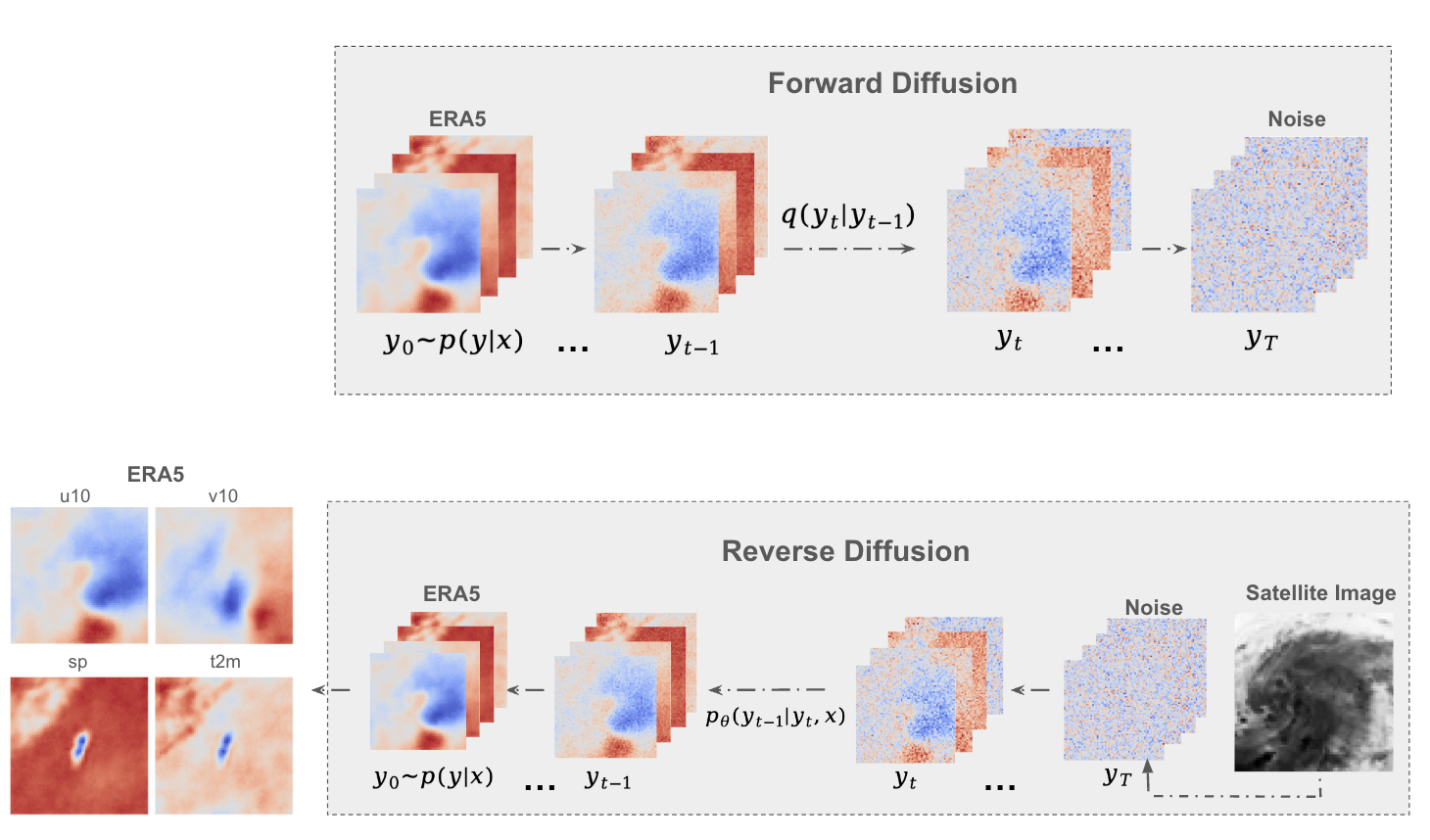}
    \caption{CDDPM workflow: \( \mathbf{y}_0 \) represents ERA5, and \( \mathbf{x} \) represents the DT satellite image. In forward diffusion, noise is added iteratively to \( \mathbf{y}_0 \). In reverse diffusion, the model denoises from \( \mathbf{y}_T \) back to \( \mathbf{y}_0 \), conditioned on \( \mathbf{x} \).}
    \label{fig: CDDPM}
\end{figure}

\subsection{Inference}
During inference, the trained model uses the DT data \( x \) as a condition to iteratively refine the noisy data \( y_T \) sampled from a normal distribution, gradually denoising it to generate clean ERA5 meteorological data. The process involves generating noisy versions of the ERA5 variables and iteratively refining these predictions by removing the noise using the learned reverse diffusion steps. The inference process can be summarised as follows:
\begin{enumerate}
    \item \textbf{Initialise with Noise}: Start with a sample of pure Gaussian noise, \( y_T \), which serves as the initial noisy input.
    \item \textbf{Iterative Denoising}: For each time step \( t \) from \( T \) to \( 1 \):
    \begin{itemize}
        \item Use the model to predict the denoised data \( y_{t-1} \) from the current noisy data \( y_t \) and the conditional input \( x \).
        \item If \( t > 1 \), sample noise \( z \sim \mathcal{N}(0, I) \); otherwise, set \( z = 0 \).
        \item Compute the next step \( y_{t-1} \) using the predicted mean and variance.
    \end{itemize}
    \item \textbf{Final Prediction}: The final output \( y_0 \) is the clean ERA5 meteorological data reconstructed from the noisy input, conditioned on the DT data.
\end{enumerate}

\section{Results}
\subsection{Magnitude Prediction}
\begin{figure}[!h]
    \centering
    \includegraphics[trim = {0.5cm 0.1cm 0.4cm 0.2cm}, clip,width=1\textwidth]{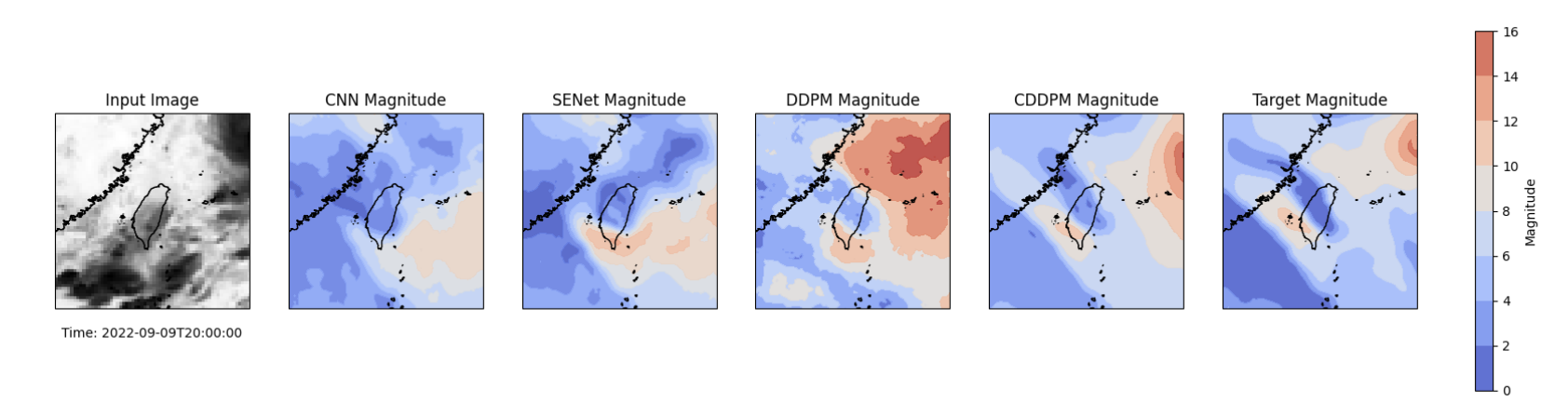}
    \caption{An example prediction of Typhoon Muifa, from September 9th 2022, showing the forecasting magnitude derived from predicted u10 and v10 components.}
    \label{fig: magnitude}
\end{figure}

The magnitude of wind $M$ ($m s^{-1}$) is a critical metric for understanding the overall wind speed and direction, and is computed as the Euclidean norm of the two wind components, u10 and v10, defined as $M = \sqrt{u_{10}^2 + v_{10}^2}$. Typhoon Muifa was a significant typhoon that impacted the Western North Pacific region in September 2022 and Fig. \ref{fig: magnitude} depicts the prediction of our models on an satellite image of this event. Clearly, CDDPM presents the closest prediction to the ground truth.

\subsection{Performance Metrics}

\begin{table}[h!]
    \centering
    \small
    \caption{Mean evaluation results over the entirety of the test dataset. Values in bold indicate the best performance.}
    \begin{tabular}{@{}p{1.2cm}p{1.2cm}p{1.3cm}p{1.2cm}p{1.2cm}p{1.2cm}p{1.2cm}p{1.2cm}p{1.2cm}@{}}
        \toprule
        \textbf{Model} & \textbf{Variable} & \textbf{KL-Div $\downarrow$} & \textbf{RMSE $\downarrow$} & \textbf{MAE $\downarrow$} & \textbf{PSNR $\uparrow$} & \textbf{SSIM $\uparrow$} & \textbf{FID $\downarrow$} & \textbf{LPIPS $\downarrow$} \\
        \midrule
        \RaggedRight CNN 
        & u10 & 0.003 & 0.039 & 0.029 & 29.441 & 0.886 & 0.068 & 146.522 \\
        & v10 & 0.004 & 0.043 & 0.032 & 28.955 & 0.872 & 0.064 & 135.013 \\
        & t2m & 0.000 & 0.024 & 0.021 & 33.529 & 0.991 & 0.004 & 27.874 \\
        & sp & 0.003 & 0.039 & 0.029 & 29.741 & 0.913 & 0.024 & 149.301 \\
        & Mean & \textbf{0.003} & 0.036 & 0.028 & 30.417 & 0.916 & 0.040 & 114.678 \\
        \midrule
        \RaggedRight SENet
        & u10 & 0.004 & 0.041 & 0.031 & 28.594 & 0.883 & 0.069 & 170.500 \\
        & v10 & 0.004 & 0.044 & 0.034 & 28.305 & 0.869 & 0.063 & 148.516 \\
        & t2m & 0.000 & 0.014 & 0.012 & 38.101 & 0.991 & 0.004 & 41.946 \\
        & sp & 0.003 & 0.041 & 0.029 & 29.404 & 0.916 & 0.025 & 135.948 \\
        & Mean & \textbf{0.003} & 0.035 & 0.026 & 31.101 & 0.915 & 0.040 & 124.227 \\
        \midrule
        \RaggedRight CDDPM 
        & u10 & 0.004 & 0.037 & 0.027 & 30.973 & 0.900 & 0.052 & 85.518 \\
        & v10 & 0.004 & 0.041 & 0.031 & 30.199 & 0.891 & 0.051 & 91.219 \\
        & t2m & 0.000 & 0.013 & 0.011 & 39.751 & 0.995 & 0.003 & 12.820 \\
        & sp & 0.003 & 0.039 & 0.028 & 30.305 & 0.929 & 0.024 & 76.502 \\
        & Mean & \textbf{0.003} & \textbf{0.032} & \textbf{0.024} & \textbf{32.807} & \textbf{0.929} & \textbf{0.032} & \textbf{66.514} \\
        \bottomrule
    \end{tabular}
    
    \label{tab:evaluation_results}
\end{table}

When considering the overall performance across all variables, CDDPM emerges as the most reliable and effective model. It consistently achieves the highest PSNR and SSIM values, along with the lowest KL-Div, RMSE, FID and LPIPS, making it the top choice for applications requiring precise and realistic predictions. SENet and CNN also found to perform well, also making it a strong candidate for applications.

\section{Conclusion}
This study evaluates the effectiveness of different machine learning models, including CNN, SENet, and CDDPM. They predict multiple meteorological variables simultaneously using satellite images, with a focus on the region surrounding Taiwan. Overall, CDDPM proved to be the most reliable for meteorological forecasting by evaluating performance metrics and magnitude prediction. 

Future work would include testing the models across different geographical regions and weather phenomena to validate their generalisability and robustness. Incorporating temporal dynamics by using time series data can improve predictions of how meteorological variables evolve, particularly in rapidly changing systems like typhoons. Additionally, exploring multimodal models by integrating diverse data types, such as radar could enhance the robustness and accuracy of predictions. 

\begin{ack}
P. Nath was supported by the \href{https://ai4er-cdt.esc.cam.ac.uk/}{UKRI Centre for Doctoral Training in Application of Artificial Intelligence to the study of Environmental Risks} [EP/S022961/1].
\end{ack}. C. Quilodrán-Casas was supported by National Center for Artificial Intelligence CENIA FB210017, Basal ANID and UKRI-IAA Purify (PSP411).

\bibliographystyle{vancouver}
\bibliography{references}
\clearpage

\appendix
\renewcommand\thesection{Appendix \Alph{section}}
\renewcommand\thesubsection{\Alph{section}.\arabic{subsection}} 
\renewcommand\thetable{\Alph{section}.\arabic{table}}  
\renewcommand\thefigure{\Alph{section}.\arabic{figure}}  
\setcounter{table}{0}
\setcounter{figure}{0}

\section{Methodology of other models}
\subsection{CNN}
\begin{figure}[!h]
    \centering
    \includegraphics[width=0.75\hsize]{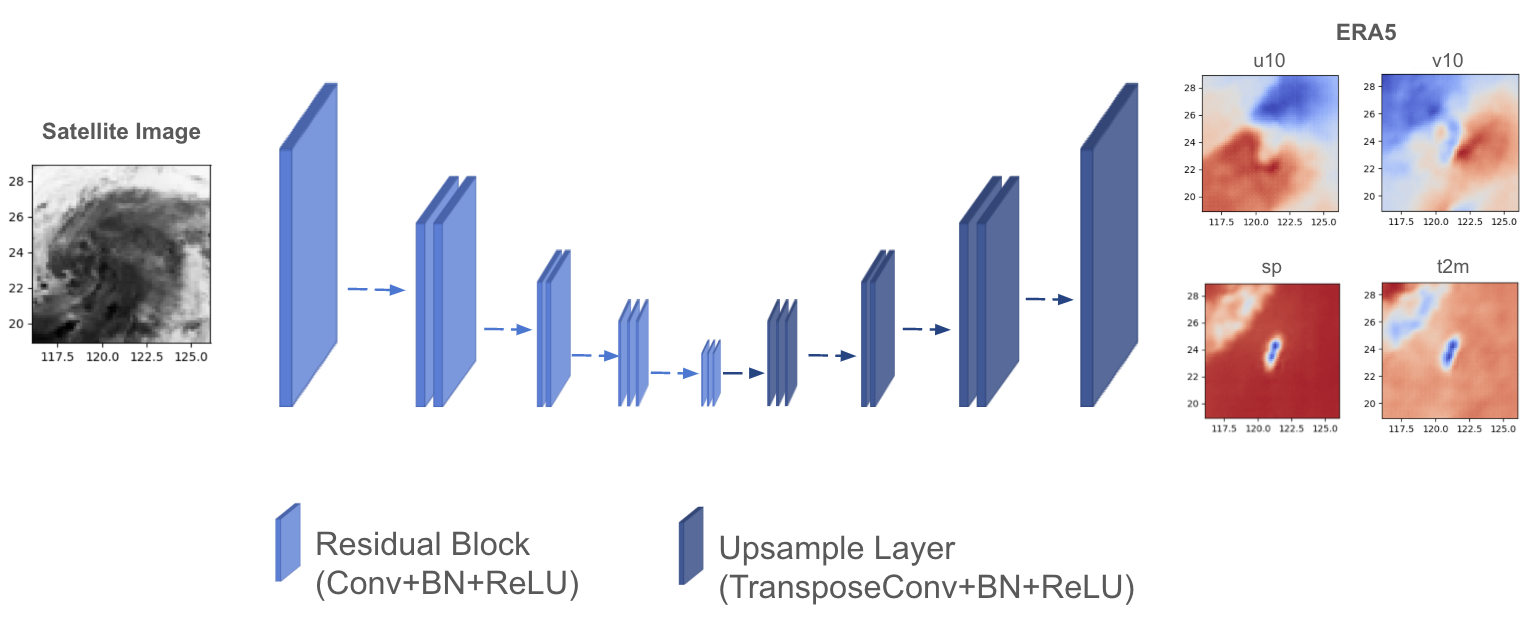}
    \caption{Structure of Conditional Convolution Neural Networks}
    \label{fig: cnn_structure}
\end{figure}
The proposed CNN architecture processes a single-channel satellite image from the Digital Typhoon dataset by replicating it into a four-channel input. This aligns the image with the four ERA5 meteorological variables used in training: u10 (zonal wind), v10 (meridional wind), sp (surface pressure), and t2m (2-meter temperature). Replicated four-channel input image is processed through the four residual blocks. Each block transforms the input feature maps while preserving the spatial dimensions. After the residual blocks, the feature maps are passed through the four image upsample layers, each of which increases the spatial dimensions until the output matches the original input size.

\subsection{SENet}
Since this project involves a multi-task scenario with four channels, each representing a distinct meteorological variable, SENet is chosen due to its ability to model the interdependencies between channels effectively. The SENet architecture builds upon the baseline CNN by integrating Squeeze-and-Excitation (SE) blocks within each residual block. It allows the network to capture more intricate channel-wise dependencies, improving the overall feature representation.




\section{Additional Results}
\subsection{Visual Result}
\begin{figure}[!h]
    \centering
    \includegraphics[width=1\hsize]{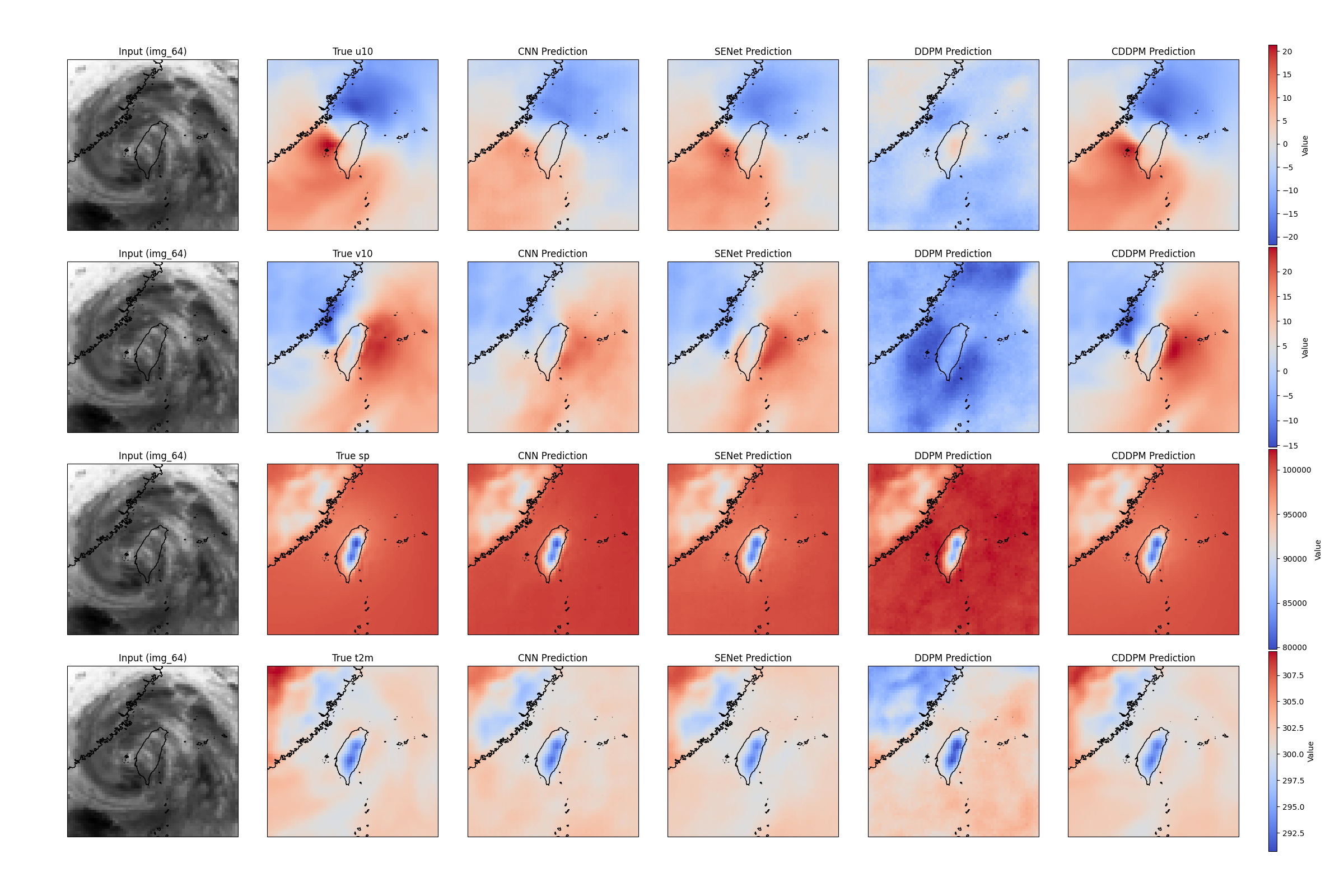}
    \caption{Example results comparing the predictions of four different models: CNN, SENet, DDPM, and CDDPM. The first column shows the input image (img\_64) used to generate predictions. The second column displays the true values of the u10, v10, sp, and t2m variables. While the subsequent columns present the corresponding predictions from each model.}

    \label{fig: results_2_17}
\end{figure}

The figure \ref {fig: results_2_17} shows example results for four different models: CNN, SENet, DDPM, and CDDPM. The predicted meteorological data for four variables were compared: u-component wind (u10), v-component wind (v10), surface pressure (sp), and temperature at 2 meters (t2m).
\subsection{Pixel Difference}
The pixel difference metric quantifies the discrepancy between the predicted and actual pixel values in the generated meteorological data.

\begin{table}[!h]
    \centering
    \caption{Mismatch Values for Different Models and Variables.}
    \begin{tabular}{@{}p{2cm}p{1.7cm}p{1.7cm}p{1.7cm}p{1.7cm}p{2cm}@{}}
        \toprule
        \textbf{Model} & \textbf{u10} & \textbf{v10} & \textbf{t2m} & \textbf{sp} & \textbf{Mean} \\
        \midrule
        CNN & 149.539 & 222.422 & 4.865 & 197.252 & 143.520 \\
        SENet & 151.089 & 234.270 & 0.887 & 214.000 & 150.061 \\
        DDPM & 1128.631 & 1829.025 & 219.908 & 568.504 & 936.517 \\
        CDDPM & 181.326 & 225.301 & 1.684 & 205.004 & 153.329 \\
        \bottomrule
    \end{tabular}
    
    \label{tab:mismatch_results}
\end{table}

Based on Table \ref{tab:mismatch_results} shown above, CNN has the lowest mismatch overall, while SENet and CDDPM also work well. However, CDDPM stands out in performance metrics but has a higher pixel difference, which can be because of small but numerous local errors, leading to a higher overall pixel difference. These discrepancies may occur because the model’s loss function prioritises minimising the noise during the reverse diffusion process. The primary goal of this process is to accurately recover the overall structure and distribution of the original data rather than focusing on precise pixel-by-pixel reconstruction.

\begin{figure}[!h]
    \centering
    \includegraphics[width=1\hsize]{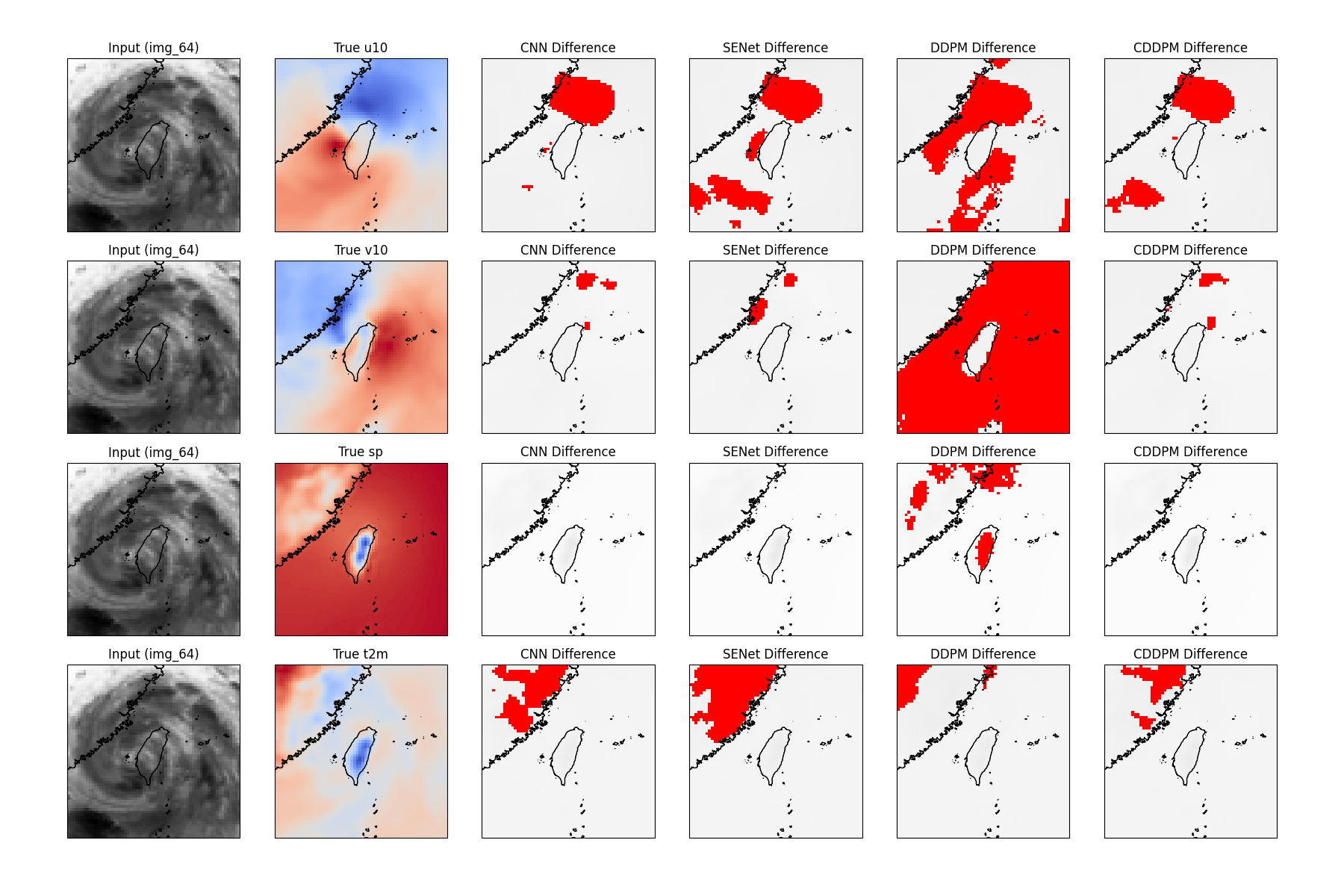}
    \caption{Difference maps for predictions of u10 (u-component wind), v10 (v-component wind), sp (surface pressure), and t2m (temperature at 2 meters) variables using four models: CNN, SENet, DDPM, and CDDPM. The first column represents the input image used for predictions. The second column shows the true values of each variable. The subsequent columns display the difference between the true values and the predicted values from each model. Red regions indicate areas of higher discrepancy between the true and predicted values, highlighting where each model deviates most from the ground truth.}
    \label{fig: difference}
\end{figure}

\end{document}